\documentclass{article}

\usepackage{amsmath}
\usepackage{natbib}
\usepackage{cleveref} 
\usepackage[ruled]{algorithm2e}
\usepackage{url}
\usepackage{color}
\usepackage{graphicx}
\usepackage[export]{adjustbox}

\begin{document}

\title{Navigating Fairness Measures and Trade-Offs}
\author{Stefan Buijsman\footnote{TU Delft, Jaffalaan 5, 2628 BX, Delft, The Netherlands, \\s.n.r.buijsman@tudelft.nl}\\ Published in \textit{AI and Ethics},\\ \url{https://link.springer.com/article/10.1007/s43681-023-00318-0}}
\date{}

\maketitle

\begin{abstract}
In order to monitor and prevent bias in AI systems we can use a wide range of (statistical) fairness measures. However, it is mathematically impossible to optimize for all of these measures at the same time. In addition, optimizing a fairness measure often greatly reduces the accuracy of the system \citep{kozodoi2022fairness}. As a result, we need a substantive theory that informs us how to make these decisions and for what reasons. I show that by using Rawls' notion of justice as fairness, we can create a basis for navigating fairness measures and the accuracy trade-off. In particular, this leads to a principled choice focusing on both the most vulnerable groups and the type of fairness measure that has the biggest impact on that group. This also helps to close part of the gap between philosophical accounts of distributive justice and the fairness literature that has been observed \citep{kuppler2021distributive} and to operationalise the value of fairness.
\end{abstract}

\section{Introduction}

One of the main risks accompanying the use of artificial intelligence in decision making is that the algorithms that are used are biased, and as a result can lead to unfair outcomes \citep{pessach2020algorithmic}. In particular, artificial intelligence is prone to (unintentionally) indirectly discriminate against certain groups. Machine learning systems (a type of AI) are fitted to data and find patterns in that data in order to predict a target variable. In doing so, they often use correlations present in the data (e.g. between ethnicity and zip codes, as with segregated neighbourhoods the zip code is a good predictor for ethnicity) to select on a problematic property (ethnicity) not directly but through the use of information on an unproblematic property (zip codes). This means that often these systems do not have direct access to variables that would be unfair to select on, but they still produce outputs that would lead to unfair treatment of certain groups. Put more precisely, indirect discrimination is the situation where a group A (e.g. people with a certain ethnicity) is treated disadvantageously not by direct reference but by disadvantageous treatment of a group B (e.g. residents of certain zip codes) with features that strongly correlate with those of group A \citep{loi2021choosing}. The complex patterns that machine learning systems can find in data are also the ones that can facilitate this kind of indirect discrimination. 

To help prevent this from happening,a range of (statistical) fairness measures has been developed \citep{carey2022statistical, mehrabi2021survey}. These measures consider the statistical distribution of outputs between groups. For example, the demographic parity measure states that ideally the probability of a positive decision (e.g. getting a loan) is equal for two groups. Using the outputs of a machine learning system it can then be determined how similar these probabilities really are, giving us a quantifiable measure of how fair the system is in the sense captured by demographic parity. Other measures such as equal opportunity focus on other probabilities. Equal opportunity states that the probability of a true positive (a correct positive decision, e.g. a loan you can afford) should be equal between two groups. Yet other measures look at equal probabilities of false positives (e.g. the odds of getting a loan you cannot afford). There are many more statistical (and indeed also non-statistical) fairness measures. 

The benefit of using these fairness measures is that they capture both direct and indirect discrimination. Moreover, they can be used to correct (at various stages of model development) biases, by enforcing that the output of the AI system comply with the mathematically specified fairness criterion. This results in fair treatment under the understanding that fits that particular measure (e.g. equal true positive rates). However, there is a wide range of statistical (and other, such as causal) measures available and by now a number of mathematical impossibility theorems \citep{kleinberg2016inherent} have shown that in virtually all cases it is mathematically impossible to optimize for all of these measures at the same time. We have to optimize, for example, either for equal probability of getting a positive decision, or for equal true positive rates, or for equal false positive rates. But we cannot have all three being equal at the same time. For example, it is mathematically impossible to make loan decisions in such a way that men and women (1) get loans equally often and (2) default on loans equally often. As long as the income distribution is not the same, one of the two groups will either get fewer loans (because the lower-income group cannot afford as many loans) or they will default more often (to get the same amount of loans as the high-income group). There is thus no way to jointly satisfy all these fairness measures, and instead we will have to choose which fairness measure to prioritize in each case.

Furthermore, the enforcement of a fairness measure in the development cycle of an algorithm reduces the accuracy of an algorithm \citep{corbett2017algorithmic}. Essentially, a fairness measure forces the system to treat two groups the same (in some way) where there are in reality underlying differences, e.g. in income distribution for a financial prediction. Enforcing equal treatment thus forces the system to ignore some of these differences, making the predictions less accurate. This leads to a further trade-off between fairness and accuracy \citep{kozodoi2022fairness}. The question, therefore, is what fairness measure to pay most attention to in a specific context and to what extent performance on this measure should be improved at the cost of accuracy, again given a specific context. 

Current approaches to deal with these two difficult choices vary. One option is to highlight a range of fairness measures along with accuracy (overall and for different sub-groups) as is suggested by the model cards framework \citep{mitchell2019model}. While this doesn't give a method to make the choices, it could be seen as advocating for the gathering of as much information on an algorithm's accuracy and fairness as one can get, to guide discussions. For specific use cases choices are certainly made between fairness metrics and on the trade-off between fairness and accuracy, but there are very few published methods for making these choices in a principled manner.

Of the published work, there seem to be three approaches. The first is to aim to settle the choice using mathematical methods, calculating a Pareto optimal solution between the different measures \citep{little2022fairness, wang2021understanding, wei2022fairness}. The second \citep{ruf2021towards} aims to structure the range of statistical measures based on a mixture of one's goals (e.g. whether one wants an affirmative action policy or not, which type of error is most sensitive to fairness) and technical features (label bias, output type, presence of explaining variables). The result is a decision tree leading to specific fairness measures. The third approach stems from philosophy and consists of work that aims to motivate the choice between the different criteria using ethical theories \citep{loi2021choosing,lee2021formalising}. These, and philosophical understandings of fairness more generally speaking, tend to be disconnected from the more technical approaches \citep{kuppler2021distributive}. 

The goal for this paper is to further develop a philosophical underpinning for choosing both what fairness metric to prioritize and for trading off between fairness and accuracy, and to begin to connect this to the more technical work on trade-offs. The current frameworks are discussed in section 2, along with arguments why we need such a philosophical underpinning for the mathematical and decision tree approach. This leads up to the discussion of Rawls' views on distributive justice in section 3, which are used to offer an alternative method for deciding between fairness metrics and the fairness-accuracy trade-off. Section 4 finally discusses limitations of the current proposal, linked to more general (and already known) limitations of statistical fairness measures. 

\section{The need for a substantive theory for fairness trade-offs}

To begin with, it is good to motivate why a philosophical approach to these trade-offs is needed and what it should provide. As decisions about which fairness measure to optimize, as well as how to balance fairness and accuracy, have serious consequences for the behaviour of the algorithm and its long-term impacts \citep{liu2018delayed} we need to know which choice to make and why. A substantive theory should therefore be action-guiding: it should present us with a clear outcome on both the fairness measure to optimize and the best trade-off between fairness and accuracy, assuming that such a best option exists. If it doesn't, then we should get a number of equally good alternatives, in a substantive sense of equally good. We should have no fairness-related reason to prefer one over the other. That shows us that we also need to know what counts as reasons for preferring one option over another, and thus that this substantive theory should give us more than a way to pick one option out of the many that are available. It should tell us why this is the best option, relating the fairness choices we make to normative considerations about fairness and justice. Without such reasons, one risks making a trade-off that is hard to defend (as one lacks normative reasons for it) and thus not accepted by society. While current approaches, which I survey next, often strive to provide the former there is still a need for a theory that has this mixture of both guiding actions and supporting us to provide reasons for those actions.

\subsection{Technical approaches: decision trees and Pareto frontiers}

\begin{figure}
\includegraphics[scale=.7, left]{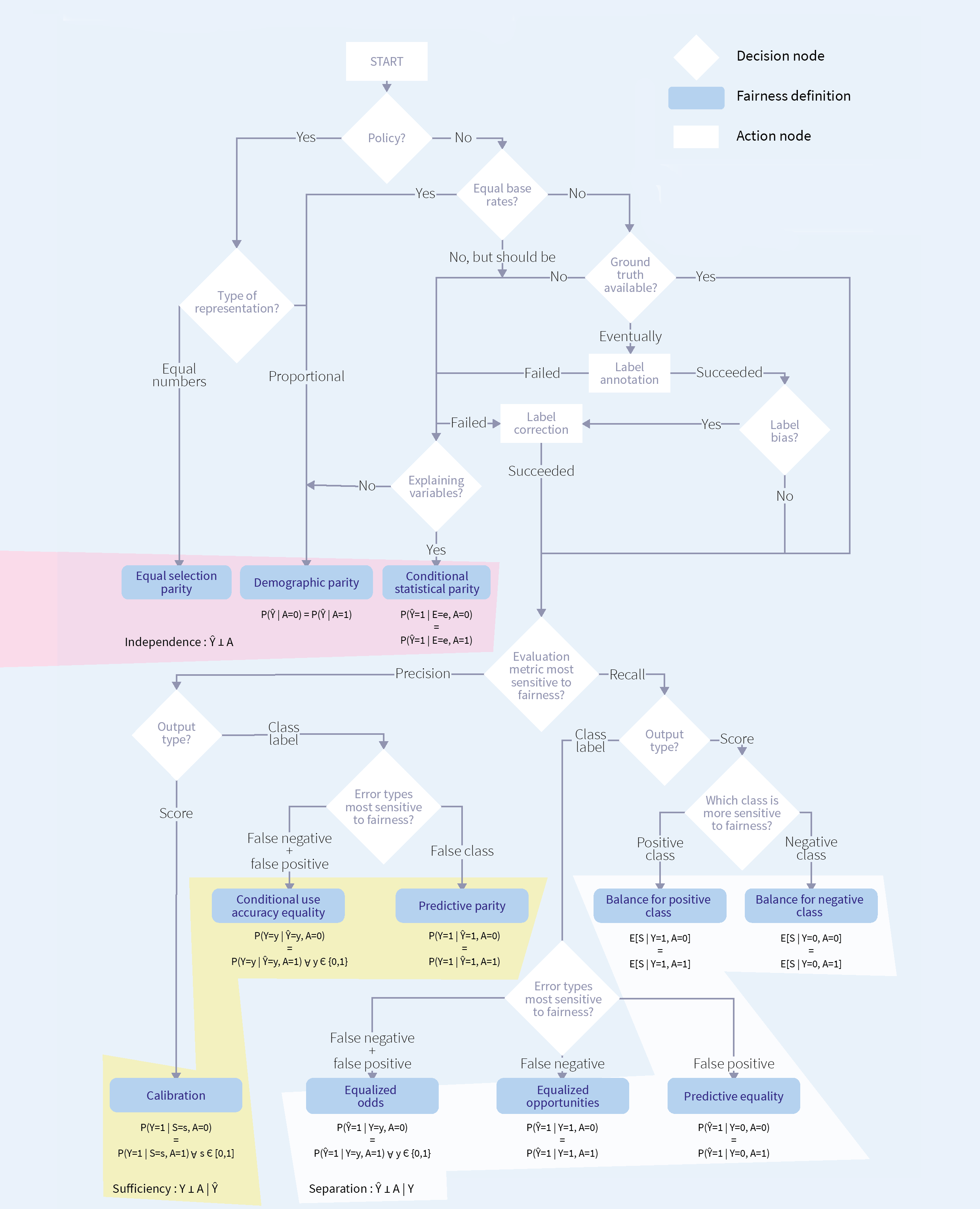}
\caption{The decision tree for picking a fairness measure taken from \cite{ruf2021towards}}
\end{figure}

To start with, the decision tree presented by \cite{ruf2021towards} and reproduced in figure 1 is a prime example of an attempt to help guide decisions on fairness measures using a mixture of technical and contextual decision nodes. It is clearly action-guiding, as it delivers a unique fairness measure depending on the choices one makes in answer to the various nodes in the decision tree. To illustrate this, it helps to consider a case such as that of a bank using an AI system to determine whether a loan should be approved. The first question to answer is then whether there is a policy to boost underprivileged groups. Should the bank wish to do so, then the next question is what kind of representation (equal numbers or proportional) is relevant. Proportional likely seems the better choice here, as we don't aim for a certain number of loans for e.g. different ethnicities, but rather want equal access to credit. The decision tree then tells us to choose demographic parity as fairness measure. We get a decision, and we have some motivation (namely, a policy to boost underprivileged groups to achieve proportional representation)

Useful as this decision tree is, it also highlights why a more substantive account for making decisions around fairness is needed. First, the choices represented by these nodes -- whether there is a need to boost underprivileged groups and whether this representation should be proportional or in equal number -- need to be motivated. This is even more so for the other decision nodes asking what features of the model are most sensitive to fairness, as this decision will depend on one's more substantive views on fairness. The authors don't dispute that, and aim rather to lay out the choices in a structured manner. Still, it shows that the decision tree can only be used properly if we have an idea on how to make these choices, and that is what we still have to provide.

Second, and more importantly, as we lack such a theory at the moment it is easy to go through these choices with good intentions but end up with a fairness measure that fails to line up with one's intentions. To see how this can happen, the path just described leads to the choice of demographic parity as the right fairness measure for the loan approval case. This makes intuitive sense, but as \cite{liu2018delayed} show in the loan approval scenario there are unintended consequences when following the advice to adopt demographic parity. In fact, optimizing for demographic parity can cause active harm for the underprivileged group that it is supposed to help, worsening their financial position in the long run and deepening financial inequality. The reason is that demographic parity can lead to a relatively higher proportion of unaffordable loans being offered to the underprivileged group, increasing (costly) defaults and thus harming their finances. Now, a reasonable response would be that the goal should not have been to boost proportional representation of the underprivileged group. Instead, we should probably have said that there is no policy to boost this group, that a ground truth is eventually available (we learn who defaults and who doesn't), succeeded in label annotation, evaluate based on recall and care most about the false positive error type. When following that path the decision tree tells us to choose predictive equality, which is the measure we need to prevent the disparate long-term impacts observed by \cite{liu2018delayed}. My point is rather that the intention to improve access to financing is one that makes sense, and that with only the decision tree at hand it can be easy to end up with a fairness measure that does more harm than good, especially considering the somewhat ad hoc way in which fairness toolkits are sometimes used in practice \citep{deng2022exploring}. Without an additional way of evaluating whether the resulting fairness measure has the effects one intended (or even stronger, should strive for) it can be hard to avoid such mistakes. Decision trees such as these can be helpful, but it is important to have a basis for evaluating the correctness of the decision tree and the way in which it has been used for a specific algorithm. 

This need is equally present for the more technical approaches to optimize trade-offs between fairness and accuracy that can be found \citep{little2022fairness, wang2021understanding, wei2022fairness}. Here, one has to choose first which fairness measures to use in the multi-objective optimization methods (and then optimizing for e.g. demographic parity in a loan approval case could still lead to a harmful situation). Second, the technical measures only calculate which situations are such that you cannot improve fairness further without sacrificing accuracy, or cannot improve accuracy without sacrificing fairness. In other words, they calculate the Pareto frontier of optimal choices between fairness and accuracy. But along that frontier there is still a difficult choice to make on what point of the Pareto frontier one picks, as options will range from maximal accuracy with a low fairness score all the way down to fully optimized fairness at a high cost to accuracy. There are no technical answers to either of these choices, and so a substantive theory is needed. The Pareto-optimization methods are still of immense help to identify our options and thus to find the best trade-off once we do have a normative criterion to decide between alternatives. However, they are not sufficient to guide our choices. The next point to consider, therefore, is to what extent we can find such a normative criterion to guide choices regarding fairness. 

\subsection{Normative approaches: Key Ethics Indicators and weighing normative frameworks}

\cite{lee2021formalising} offer a six-step procedure to formalise trade-offs, and in particular those around algorithmic fairness. The first step is to define ``ethical success" for the algorithm (benefits and harms, affect on fundamental rights). Next, the layers of inequality (step 2) and bias (step 3) should be identified. After this, mitigation strategies should be formulated (often extending beyond the use of fairness measures) in step 4. The fifth step is to operationalize the different objectives from step 1-3 and calculate/visualize the trade-offs between different options. Finally, one should select the model that best fits one's values and objectives. As becomes clear from this series of steps, the main purpose of this procedure is to offer a stronger basis for making the trade-offs and to advocate for the inclusion of a broader set of (ethical) goals that accompany the design and implementation of an algorithm. This can shift perspectives, as they (for example) ``operationalise financial inclusion as the total expected value of loans under each model and minority loan access as the loan denial rate of black applicants under each model" \cite[p.541]{lee2021formalising}. While this particular trade-off can also be captured in terms of accuracy and fairness measures, the point is that inequality is a broader issue that can link to more than the output distribution of the algorithm. That being said, relatively little guidance is offered on step 6, choosing amongst the now quantified objectives. 

Instead, in support of step 6, the authors give an overview of philosophical perspectives on inequality, highlighting the differences of opinion that are prevalent there (see also \cite{kuppler2021distributive}) but without choosing sides or substantially saying what choice would be made on what framework. They also discuss a number of perspectives from welfare economics, in particular on beneficence (ensuring that everyone is as well off as can be), non-maleficence (avoiding harm as in the loan case discussed earlier), autonomy (ensuring there are sufficiently many quality options for people to freely choose from) and explicability. These considerations, such as whether there is a double-standard in privacy between those whose credit scores are calculated using their bank information and those whose credit scores are calculated using alternative data such as browser history, are good to include in the decision making process regarding trade-offs. However, no clear standard emerges from this discussion on how to weigh the different concerns. The choices are clarified, and some reasons that can be used to make these choices are offered (getting us closer to the normative side of what we want), but they are not sufficiently action-guiding to tell us what trade-off to make between fairness and accuracy in a specific context.

In contrast, \cite{loi2021choosing} focus on developing a method to directly rank different alternatives given their fairness and accuracy. Their approach is similarly based on an overview of different philosophical positions on inequality, using a total of six different moral principles to score alternatives. These are (1) discrimination on features not subject to choice (e.g. gender), (2) discrimination on features that depend on other people's choice (e.g. the colour of a car, which may correlate with accident rates), (3) discrimination on unjust statistical facts, (4) utilitarianism, (5) prioritarianism and (6) egalitarianism. Different algorithms are given (normalized) scores on these six measures. For example, the algorithm that does worst on indirect discrimination on gender is given a score of -100, whereas less discriminatory algorithms score -50 or (in an ideal case) 0. Likewise, a more accurate algorithm may score higher on utilitarianism (which focuses on the maximum total utility) than a less accurate one, under the assumption that e.g. more loans will be provided when using a more accurate algorithm. To make a final choice between algorithms one has to give probability estimates of how likely it is that the different moral principles are true, allowing for a weighted sum of the scores. Or one has to settle for a consensus-based approach in which case only some trade-offs can be solved (as there are going to be cases on which the different perspectives fail to agree on a single choice). 

There is a clear merit to this kind of approach, namely that it considers the full spectrum of ethical perspectives. At the same time, it illustrates that this broad approach to offering a substantive backing for fairness choices is unsatisfactory. There are no clear guidelines on how to determine scores for algorithms on the six criteria, nor is there a principled manner to determine the weights given to these criteria. In fact, the choice between the overarching criteria may be harder to make as it is a much more general, abstract choice between moral principles as opposed to the choice to e.g. prioritize false positives over false negatives in a specific use case. As a result, this method does not seem very promising to help tackle the different trade-offs that need to be made in the development and implementation of an algorithm. 

As a conclusion of this brief overview of current approaches to making trade-offs related to fairness, the main take-away is that there are currently very few aids to make these substantive choices. Yes, there are technical tools and decision trees that help to identify the alternatives and to structure the discussion, but there is little help in what counts as a good reason for making the choice one way rather than another. While there are some discussions in philosophy that aim to support these choices, they tend to opt for a very wide approach, including all the different philosophical perspectives. The consequence is that the guidance remains vague, and difficult to use in practice for developers. To get closer to concrete guidance on fairness decisions I therefore opt to go the other route. In the next section I zoom in on a specific account of distributive justice, by John Rawls \cite{rawls2001justice}, and aim to work out the implications of that for choosing fairness measures as well as deciding on the fairness--accuracy trade-off. This will only be the first step in better support for decision making around fairness and would ideally be complemented by similar efforts for other philosophical perspectives. Still, it is a step in the right direction.

\section{Rawlsian distributive justice and fairness}
One of the most influential accounts of distributive justice (i.e. accounts that specify how benefits and harms should be distributed in society) is that by \cite{rawls2001justice}. This account consists of two parts, one requiring an equal distribution of basic liberties (and in earlier formulations, rights). The second, and better known part, stating that inequalities are allowed as long as they result in benefits for the worst off (also known as the Difference Principle). In his words this means that there are two principles of justice:

\begin{quote}
    1. Each person has the same indefeasible claim to a fully adequate scheme of equal basic liberties, which scheme is compatible with the same scheme of liberties for all; and \\
    2. Social and economic inequalities are to satisfy two conditions: first, they are to be attached to offices and positions open to all under conditions of fair equality of opportunity; and second, they are to be to the greatest benefit of the least-advantaged members of society. \cite[p.42-43]{rawls2001justice}
\end{quote}

In cases of conflicts, principle (1) takes precedence over principle (2). Furthermore, the first part of principle (2), on equality of opportunity, again takes precedence over the second part, that allows inequalities as long as they benefit the worst-off. Rawls thus presents us with the following picture: we should strive for equality of opportunity, such as equal access to education and healthcare, to provide as level a playing field as we can. However, some inequalities are permissible, as long as these inequalities are of net benefit to the worst-off. That idea is rooted in the observation that wealth creation (for everyone, e.g. through technological advancement) benefits from the presence of some inequalities. Rewards may be used to stimulate innovation, as long as one makes sure that the benefits of innovation are not restricted to the fortunate members of society.

This makes Rawls' approach to fairness different from much of the current algorithmic fairness literature. His difference principle focuses on the absolute level of welfare, rather than at relative differences between groups. In addition, his concern is (for this principle) not explicitly for attributes that we do not choose such as gender and race, but rather for one's overall welfare level. It's the worst-off that society should care for, at the end of the day. Now, one shouldn't forget the earlier principles advocating for equality of opportunity and equal basic liberties, which do talk about relative equality for all members of society. Note, however, that Rawls does not mean equality of opportunity in the sense of the fairness measure of the same name (which entails equal true positive rates for two groups). Rather, the philosophical use of the term refers to the idea that these positions should be open to all and that one gets these positions on merit, rather than on `lucky' attributes such as race, gender, or even natural talent. There is a wide debate on how substantive this equality of opportunity should be, i.e. how much opportunity provision is needed to correct for natural and societal inequalities \citep{sep-equal-opportunity}. It thus encompasses much more than the algorithmic fairness notion, and focuses more on the provisions in society to offer everyone equal opportunities to live healthy, fulfilling lives. This certainly can translate into fairness measures, and would relate to the policy question in the decision tree discussed in section 2, but is not meant to govern every (algorithmic) decision we make.

Before diving in to the implications of Rawls' views for the choice of fairness metric and the trade-off between fairness and accuracy, there is one last general point to clear up. Rawls intends for his account to describe how a just society would function. On the other hand, most of the algorithms that are discussed/developed do not have such a broad scope. Companies therefore may not feel the need to adhere to these principles, and as a matter of fact do not tend to focus on ensuring that their wealth-creation also benefits the worst-off. At the same time, companies are also a part of society and their actions have serious impacts on how society is structured. If we want to realize a society where these principles are adhered to, then it cannot be that only governments follow them. Equality of opportunity and a concern for the welfare of the worst-off should be widely shared in such a case, governing decisions on the distribution of benefits and harms regardless of who makes them. Therefore these principles should be applicable also at a smaller scale than for the entire society. It may be that for pragmatic reasons one cannot fully live up to them, and one may not expect that a company also actively helps tackle inequalities in access to education (though it would be laudable), but the mere fact that Rawls had a broader scope in mind shouldn't prevent us from applying them at a smaller scale. Assuming that this is fine, what would follow?

\subsection{The difference principle as a guide in trade-offs}

As there are more difficulties in the interpretation of equality of opportunity, and the scope of that principle is intended to be smaller, it makes sense to start with the application of the difference principle. We should then make choices based on their impact on the absolute level of welfare of the worst-off. The challenge then is to somehow capture this level of welfare as well as to identify which group is meant by the worst-off. While it may not be possible to pin this down perfectly, it need not be a mysterious affair once you work with a concrete scenario. To start with, consider the loan approval case mentioned earlier, as there are a few simulation studies available for this case measuring the financial impact of different choices \citep{kozodoi2022fairness, liu2018delayed, zou2022ai}.

An interesting point to start with is the analysis by \cite{zou2022ai}. They found that when training a classifier on mortgage application data it indeed displays bias against black applicants, and in fact amplified that bias compared to the inequalities present in the dataset alone. Using fair machine learning techniques they then trained a classifier that was substantially less biased (selection rate difference was reduced from 17.4\% to 1.46\% and total approved applications increased by 165\% for black applicants compared to 6\% for other applicants). In doing so, however, the overall false positive rate went up by 362\% and the average loan amount for black applicants decreased by 36\%. As a result, individual applicants were worse off, though the group as a whole received substantially more financing (4722 applicants at \$258.280 each v.s. 12494 applicants at \$163.820 each). Furthermore, as application and approval rates do not differ much for other applicants, it is likely that a substantial number (not reported, but possible more than half) of the black applicants end up in debt because the approval counts as a false positive. Without this last bit of information it's hard to estimate the exact impact on the welfare of black applicants, but it seems plausible that, as in a substantial number of simulations run by \cite{liu2018delayed}, there was a net negative effect on their financial position as a result of the much higher false positive rate. 

It is that net negative effect, and not the lowering of individual loan amounts that the authors point to, which would be problematic on the difference principle. While the group of all black applicants may not precisely coincide with the worst-off, it is certainly a group that will be worse off compared to the average of all other applicants. For that reason, the absolute effect on their welfare should be considered. If the false positives do not weigh down the number of additional approved loans, then despite the fact that the rest of the loan applicants also received less money per application the fairness-aware model is the better one. It's the different distribution of wealth, which also happens to involve a larger total amount of loans (the other applicants get \$740 million less in loans from the fairness-aware model, which is easily compensated by the \$1273 million increase in loans to black applicants), that drives this choice. If that distribution leaves black applicants better off overall, then the decrease in accuracy of the fairness-aware model is worth it. This shows that the Key Ethics Indicators suggested by Lee et al \cite{lee2021formalising} for a similar lending scenario (namely total value of approved loans and approval percentage for black applicants) are the wrong ones on this framework. They don't inform us of the precise impact the algorithm has on black applicants, and could easily lead us into a situation where a large number of loans are approved (in total and for black applicants), but with so many false positives that everyone ends up worse off. Better, then, to look at the net effect on black applicants, in terms of both correctly and incorrectly approved loans. As long as that balance is better than that of the biased algorithm, we should choose the unbiased AI system. Provided, that is, that the resulting situation does not impact profits so much \citep{kozodoi2022fairness} that the bank will go broke in the long run. It is the long-term welfare of the group that matters, and it seems plausible that this is not helped by banks adopting loan approval policies that are financially unsustainable. 

This brief case illustrates how the difference principle can be used to make choices between fairness and accuracy. If the improvement of a fairness measure for an underprivileged group\footnote{Strictly speaking, this is the group of the worst-off according to Rawls. However, it might be worth broadening this in practice as Rawls focuses on society as a whole and this might obscure relevant groups in more specific cases. Either way, it is an important decision which groups we consider here as underprivileged and thus whose welfare we monitor for our fairness decisions. This also touches on debates on intersectionality, as a reviewer rightly pointed out.} is to their net benefit, then it is morally just (according to Rawls) to accept the accompanying decrease in accuracy. If the additional mistakes ultimately leave them worse off, then it is better to have a more accurate model. The difference principle can, at the same time, also help to steer the choice of fairness measure. As has become clear from the simulation studies false positives have the highest impact, both directly on the welfare of the individuals and on the profits of the banks. So, a fairness measure that aims to minimize the number of false positives seems like the best place to start. It may not provide the globally optimal solution, as one might minimize false positives by drastically reducing the number of approved loans, but it highlights why there is a strong tendency to think that equal false positive rates is the right metric to track here. It's the one that has the most impact on decision subjects' welfare, and that is what we ultimately care about if we follow the difference principle. 

Before considering a second case to further illustrate how the difference principle can work in practice, it is worthwhile to point out one further implication for algorithmic fairness: the group of people of low socio-economic status is also relevant to consider. This group is, in many respects, among the worst-off in society and therefore merits additional consideration when designing AI systems. If their situation is made worse by the implementation of an algorithm, then that is morally relevant and unfair in a substantive sense of the word. While it is understandable that this group hasn't been actively considered, given the roots of the fairness literature in non-discrimination concerns, normative accounts like that of Rawls highlight that we shouldn't forget about them. So, even though there are likely overlaps between the groups considered in the fairness literature and the group of people with low socio-economic status, it would be good to additionally consider them as an important group in its own right. This means that the take-aways for now are:
\begin{itemize}
    \item Focus on the fairness measure that is associated with the biggest impact on decision subjects' welfare
    \item Choices on trade-offs between fairness measures as well as between fairness and accuracy are made based on what maximizes the welfare of the worst-off
    \item As a result, it is important to look not just at gender, race, nationality and other protected attributes, but to also explicitly consider impacts on people of low socio-economic status
\end{itemize}

Having said that, a second case study looking at healthcare helps see how this can vary from situation to situation. Consider for this case an algorithm making medical diagnoses, for example for lung cancer. To choose a fairness measure to focus on the question here is (again) which outcome has the bigger impact on the welfare of patients. Naturally, a true positive has (or at least should have) a big positive effect on patients' welfare. The more difficult question is whether a false positive or a false negative has a bigger negative impact on their welfare. Ideally claims on this would be backed up by empirical data (possibly using a quantitative measure of health such as QALYs \citep{whitehead2010health}), but for now speculation will suffice to illustrate the method. False positives lead to a risk of getting impactful treatments (e.g. chemotherapy) that fail to help, or perhaps even hinder, recovery from the actual disease the patient is suffering from. Ideally this risk is mitigated by the use of independent methods to verify the diagnosis, especially before starting on major interventions. False negatives, on the other hand, risk that a diagnosis is ruled out without further tests and that a patient has to go through (many) more cycles of tests before the actual disease is identified. Depending on the disease, this lost time can be costly. 

While the weighting of the two risks will differ depending on the disease (it is much more costly for the patient to be late in identifying aggressive forms of cancer than it is to delay identification of less serious or more slowly developing conditions) and on value choices in healthcare, my expectation is that false negatives will have a bigger effect than false positives -- at least as long as the AI prediction is not the last word before treatment starts. This can then guide our selection of a fairness measure, aiming to equalize false negative rates (and thus also true positive rates) for different groups. Here one does get the possibly counter-intuitive conclusion that a (somewhat) lower accuracy -- i.e. fewer total correct diagnoses -- may be justified if the group that is worst-off receives better treatment as a result. On the other hand, an algorithm that is 99\% accurate for the well-off and at chance for the worst-off wouldn't be fair or acceptable, even if it were to have a larger number of correct diagnoses. The difference principle is one basis to make this choice, but it doesn't mean that the choice always becomes easier. It's merely that we now have a principled way of making these decisions, both between fairness measures to prioritize and in the trade-off between fairness and accuracy.

There is one last question that needs answering, namely who the worst-off are in this situation. There are at least two options: (1) the people in the worst state of health and (2) the people of low socio-economic status/the worst-off in society generally speaking. There is going to be some overlap between these groups, of course, but they are clearly distinct. The choice between which to prioritize also matters: it guides whose welfare is the priority for a fair healthcare system. I think there is something to say for both, but that ultimately the second option is the better fit with the difference principle. The principle is intended to cover a just society, and so concerns itself with the worst-off overall. That doesn't mean that those in the worst state of health should not be a priority in treatment -- a policy that focuses on treating those who need it the most first is likely to be beneficial to the societally worst-off just as it is beneficial for everyone else -- but it does mean that their welfare is not the ultimate arbiter of the fairness of our allocation of healthcare services. 

Hopefully this illustrates sufficiently how the difference principle operates. It shifts the questions of relative fairness to one of absolute difference in the welfare of those who are worst-off. While that is difficult to measure, especially in terms of the effects on the long term, it gives us a better handle than the decision criteria that are currently available (section 2). The difficulties in actually using the difference principle are explored further in section 4. First, however, there is the other part of Rawls' account to be considered: the fair equality of opportunity that should hold for the offices to which inequalities (sanctioned by the difference principle) attach. 

\subsection{Equality of opportunity}

Not all decisions should be based on the difference principle. For goods associated with one's direct participation in society, as well as access to the inequalities that may arise in society, there should be a level playing field. Hiring and education are the natural focal points for this level playing field, but the exact extent of it is still a matter of philosophical debate. It is thus, partly, up to organizations to determine if they consider a certain decision to be fundamental to people's access to society or if it is a decision where inequalities are permissible. And so, while I discussed healthcare under the difference principle, one may also argue that health is such a basic part of one's ability to take up positions in society that we should strive for equality here. With differences in health come differences in opportunity, and so it could be covered by the requirement of fair equality of opportunity. If so, what happens then?

To reiterate, equality of opportunity involves more than what is typically captured by algorithmic fairness. It also means providing people the support they need to take advantage of opportunities, for example by providing information in multiple languages or offering subsidised tutoring services to underprivileged pupils. \cite{rawls2001justice} discusses fair equality of opportunity largely in those terms; it implies that one's socio-economic status should have no impact on one's competitive prospects \citep{sep-equal-opportunity}. For that to be the case, we need to arrange a society where support is available for those who are born into a part of the population that is relatively worse off. That being said, there are of course also implications for algorithmic fairness. 

If we stick to the idea as found directly in Rawls, a causal approach to fairness \citep{kusner2017counterfactual, makhlouf2020survey} focusing on whether there is a causal impact of socio-economic status on outcomes (e.g. selection for a certain educational program) is a natural fit. This captures the idea that socio-economic status should have no impact on one's chances for access to various positions in society, but that e.g. natural talents are allowed to lead to differences. As is well-known, however, this presents difficult practical problems as we do not have causal models that accurately capture the complexity of many of these situations \citep{fawkes2022selection}. Without these causal models, counterfactual/causal approaches to fairness cannot be used. Furthermore, using causal models that incorrectly capture the actual causal links leads to wrong estimations of the impact of the features of interest on the algorithm's output, and such mistakes are hard to spot.

The simpler alternative, also found in the decision tree by \cite{ruf2021towards} when aiming for meritocratic policies, would be to opt for a (conditional) demographic parity approach instead. What we want in this case is the equal participation across a number of protected attributes, because talent is distributed equally as well. Recent work on counterfactual fairness suggests that this may be a particularly viable alternative, as algorithms that satisfy counterfactual fairness constraints also satisfy demographic parity and vice versa, modulo some modifications to the algorithm \citep{rosenblatt2022counterfactual}. The more substantive notion of equality of opportunity thus seems to coincide not with the fairness measure of the same name (equal true positive / false negative rates), but rather with the idea that if we've ensured that opportunities are truly equal then (conditional) demographic parity should hold. In the meantime (with unequal base rates), enforcing (conditional) demographic parity may be the best way to ensure that opportunities become more equal. This comes at the cost of accuracy of the decisions, and so may also come at the cost of the welfare of a society. This cost, especially when it is one that disadvantaged groups have to bear, should be carefully considered. The Rawlsian principles only require fair equality of opportunity for one's access to positions in society, and not for the distribution of all goods. For these more fundamental parts, (conditional) demographic parity and its associated costs makes sense (noting again that the best solution may still be to mitigate the situation in other ways than by enforcing a fairness measure in the algorithm). In other cases, it is better to consider what is of most benefit to the worst off.

This is in line with the much more extensive discussion on (conditional) demographic parity by \cite{foulds2020parity}. They cover a wide range of arguments for and against parity-based fairness measures and likewise argue that when (conditional) demographic parity is used on to mitigate unfair disparities it makes for a more meritocratic decision procedure. They also cover a range of situations (e.g. when a disparity is due to legitimate reasons) when parity should not be used. Added to that list can be the situation when decisions do not cover access to offices and positions in society to which further benefits accrue. 

With that, there is a more concrete guide to navigate the choices between fairness measures as well as the trade-off between fairness and accuracy. If we are in a situation where fair equality of opportunity is crucial, then partiy-based measures are to be preferred and a trade-off with accuracy should be permissible. While there has not been an explicit discussion of what trade-offs are permissible, here the focus should be on promoting access. However, one could again use the absolute impact on the welfare of the group whose access is intended to improve to guide tough choices that might arise. In that sense, the situation is clearer when the difference principle is the main guide. Then, the choice of (primary) fairness measure is based on what aspect of the algorithms' performance has the biggest impact on the welfare of the worst-off. Choices between fairness and accuracy are again guided by whatever specific choice leaves the worst-off with the highest absolute level of welfare. That is a good start, but of course this is not without its limitations when it comes to practically using it. These are discussed in the next section.     

\section{Limitations}

The main issue in applying the difference principle is that we often lack the information that is necessary to use it. We don't know exactly how well off the worst-off are, either in the current situation or (especially) in the new situation where an algorithm is used. Estimating these impacts is difficult and will likely remain difficult, as it is virtually impossible to anticipate during development how exactly an algorithm will perform and be used once it is deployed. Added to this is the further difficulty that Rawls' difference principle is focused on the long-term impact on welfare. So, if this group is initially worse off but in the long run ends up in the best possible position (in absolute terms), then that would still be the fair/right choice. Perhaps we can anticipate some of these situations, but there is little to no chance that we will do so systematically. Does this mean that, in the end, we cannot really use the difference principle?

I don't think it does. It shows that we are likely going to be uncertain about what the right choice is, and that we are going to make mistakes because we incorrectly estimated the impact of an algorithm. However, with the difference principle we have a clear measure of what matters in the estimation and evaluation of an algorithm's impact. We can then make a well-motivated choice based on our best estimate of the situation. In my opinion that choice also matches one of the main motivations behind the fairness literature: the aim to prevent negative impacts of algorithms on vulnerable groups. This aim is aligned with, and further substantiated by, the difference principle. 

It does mean that we need to shift part of the emphasis of the algorithmic fairness literature. Currently the main research focus is on comparative fairness, even found in a recent attempt to use Rawls' difference principle as the basis for a statistical fairness measure: ``(MinMax fairness) A model is fair if it does not make more systematic errors for any sub-group in the dataset compared to the others." \cite[p.8]{barsotti2022minmax} This fails to capture the focus on absolute welfare, nor does it single out the people who are worst-off in society as a whole. We thus need more research on the impacts of different fairness choices in those absolute terms. The simulation studies used in section 3.1 have shown that we can do good research of this kind, and that these can help to clarify the results our choices even if the setting is idealized. Studies of this kind can then act as a way to benchmark new developments in the fairness literature (new types of fairness-aware machine learning, new fairness measures, etc.) in terms of how they help improve the impacts that algorithms have. 

This points to a second difficulty and practical limitation of using the difference principle: there is no clear measure of welfare that can be used across use cases. While there are of course many different types of measures for welfare and (subjective) well-being \citep{diener2009new, osberg2005should, quiggin2012generalized, topp20155} these may not be applicable across all decisions one can make with an algorithm, nor can impacts on the more subjective measures be anticipated as easily as those on more quantitative measures. This is certainly a difficulty, and to apply the difference principle we would ideally have a measure that is both practical and universally applicable. The choice of welfare measure is, moreover, a very loaded one. How we measure e.g. health is informed by value choices, just as how we measure welfare will be. These choices need a similarly substantial argumentation that tells us why we look at a particular measure of welfare rather than another one. However, this doesn't mean that we cannot use the difference principle until then. The distribution of goods that are associated with the algorithmic decision (loans, health, etc.) are a good and non-arbitrary alternative, as they show the direct impact of the algorithm on decision subjects. More detailed discussions can then be had for these cases on what we are measuring (e.g. for health) and why. There may also be further-reaching consequences, as e.g. money allows people to improve their well-being in other ways, but the immediate effects of our decisions are a good place to start.

These two problems do not show up as prominently for the fair equality of opportunity principle. Here, absolute measures are not mentioned, as the point is that everyone should be given the same opportunities to develop their talents and access (desirable) positions in society. Comparative, and primarily parity-based, measures are a good fit for that principle. The main limitation here is that fair equality of opportunity covers much more than what the statistical fairness measures can tackle (as does, for that matter, the difference principle). The reason to bring this up again is that I believe that in practice it will be important to not steer blindly on parity-based measures when one's ambition is to achieve fair equality of opportunity. The absolute levels of welfare and one's impact on those should also matter, precisely because the scope of statistical measures is limited. If a parity-based algorithm ends up widening the gap between the worst-off and the rest of society (as they can in at least the loan scenario modeled by \cite{liu2018delayed}) then a parity-based approach does not manage to realize more equal opportunities. This brings us back to the earlier two limitations, and the same suggestions for dealing with them. The goal here should be a substantive, rather than formal, equality of opportunity and therefore we have to consider once again the absolute rather than relative impacts on people.

Finally, I want to stress again that I have focused here on a specific account of distributive justice. As was explicit in other choice frameworks \citep{loi2021choosing,lee2021formalising} there are more philosophical/normative accounts describing what fair distributions look like. So, while I hope that the work here helps to make it possible to make better-informed choices in the trade-offs between fairness measures and between fairness and accuracy, it should not be the ultimate guide to these choices. Other accounts would lead to other ways of making these trade-offs, and it would be good to have these spelled out as well. There is no room to do that here, but it is a worthwhile endeavour both to clarify the differences between the philosophical accounts and to allow for even more nuanced decisions on fairness. 

\section{Conclusion}

How to choose between fairness measures and what trade-off between fairness and accuracy should we make? As discussed, the approaches published so far are of limited help. Technical solutions such as calculating the Pareto frontier between fairness and accuracy can help to identify the best candidates, but need additional guidance in the selection of fairness measure and the choice between non-dominated alternatives. Decision trees can help, but leave the substantial choices open in the decision nodes. Recent philosophical work has also kept open the choices, pointing to Key Ethics Indicators for which the choice problem reiterates or re-framing the choices in a particularly abstract format with scores and weights for different ethical theories. Instead, I have suggested to take a single influential ethical framework and work out in more detail how one makes fairness choices under that framework (and encourage that the same is done for other accounts). This results in the following guidelines:

\begin{itemize}
    \item If the decisions govern people's access to offices and positions in society, then fair equality of opportunity holds. This is best captured by parity-based measures (typically conditional demographic parity) but in the algorithmic case should be verified by estimating the resulting difference in absolute welfare. The ultimate goal is substantive equality, not formal parity.
    \item In all other cases, the difference principle implies that we should choose such that the absolute welfare of the worst-off is maximized. This entails:
    \begin{itemize}
        \item Focusing on the fairness measure that has the most impact on absolute welfare, e.g. false positives in the loan case due to the high cost of default
        \item Trading off between fairness and accuracy to maximize the benefits to the worst-off
    \end{itemize}
\end{itemize}

An upshot of these guidelines is that we should pay more attention to (and do more research on) the absolute impacts of algorithms. This should include both the vulnerable groups that are currently the subject of the fairness literature and people of low socio-economic status that are often left out of these discussions. And while we may not be able to measure their welfare in a widely-agreed upon, uniform, way, we can start with measuring impacts in terms of the benefits/goods that are distributed by the relevant algorithms. I hope that then we can make more substantially motivated choices between fairness measures and in the fairness-accuracy trade-off. 

\bibliographystyle{ACM-Reference-Format}
\bibliography{sample-base}

\end{document}